\documentclass[10pt,twocolumn,letterpaper]{article}

\usepackage{iccv}
\usepackage{times}
\usepackage{epsfig}
\usepackage{graphicx}
\usepackage{amsmath}
\usepackage{amssymb}
\usepackage{subfloat}

\usepackage{booktabs}
\usepackage[dvipsnames]{xcolor}         
\usepackage{colortbl}

\usepackage{subcaption}

\newlength\savewidth\newcommand\shline{\noalign{\global\savewidth\arrayrulewidth
  \global\arrayrulewidth 1pt}\hline\noalign{\global\arrayrulewidth\savewidth}}

\newcommand{\smalltablestyle}[2]{\setlength{\tabcolsep}{#1}\renewcommand{\arraystretch}{#2}\centering\small}

\newcolumntype{*}{>{}}

\definecolor{lightgray}{gray}{0.95}

\definecolor{baselinecolor}{gray}{.9}
\newcommand{\baseline}[1]{\cellcolor{baselinecolor}{#1}}
\definecolor{Highlight}{HTML}{eaf7eb}  

\definecolor{citecolor}{RGB}{65,130,164}
\definecolor{linkcolor}{RGB}{166,64,54}
\definecolor{citecolor2}{HTML}{0071bc}
\usepackage[pagebackref=true,breaklinks=true,colorlinks,
citecolor=citecolor,linkcolor=linkcolor,bookmarks=false]{hyperref}

\iccvfinalcopy 

\def\iccvPaperID{7206} 

\ificcvfinal\fi

\begin{document}

\title{A Range-Null Space Decomposition Approach for Fast and Flexible Spectral Compressive Imaging}

\author{
Junyu Wang$^{1}$\thanks{Equal contribution. $^\dag$Corresponding author: xgwang@hust.edu.cn.} \
    Shijie Wang$^{1*}$ \
    Ruijie Zhang$^{1}$ \
    Zengqiang Zheng$^{2}$ \
    Wenyu Liu$^{1}$ \
    Xinggang Wang$^{1\dag}$
    \\
    $^{1}$ Huazhong University of Science \& Technology
    \\
    $^{2}$ Wuhan Jingce Electronic Group
}

\maketitle
\ificcvfinal\thispagestyle{empty}\fi

\begin{abstract}
We present RND-SCI, a novel framework for compressive hyperspectral image (HSI) reconstruction. Our framework decomposes the reconstructed object into range-space and null-space components, where the range-space part ensures the solution conforms to the compression process, and the null-space term introduces a deep HSI prior to constraining the output to have satisfactory properties. RND-SCI is not only simple in design with strong interpretability but also can be easily adapted to various HSI reconstruction networks, improving the quality of HSIs with minimal computational overhead. RND-SCI significantly boosts the performance of HSI reconstruction networks in retraining, fine-tuning or plugging into a pre-trained off-the-shelf model. Based on the framework and SAUNet~\cite{wang2023simple}, we design an extremely fast HSI reconstruction network, RND-SAUNet, which achieves an astounding 91 frames per second while maintaining superior reconstruction accuracy compared to other less time-consuming methods. Code and models are available at \url{https://github.com/hustvl/RND-SCI}.
\end{abstract}

\section{Introduction}
\begin{figure}[t]
    \centering
    \includegraphics[width=\linewidth,scale=1.00]{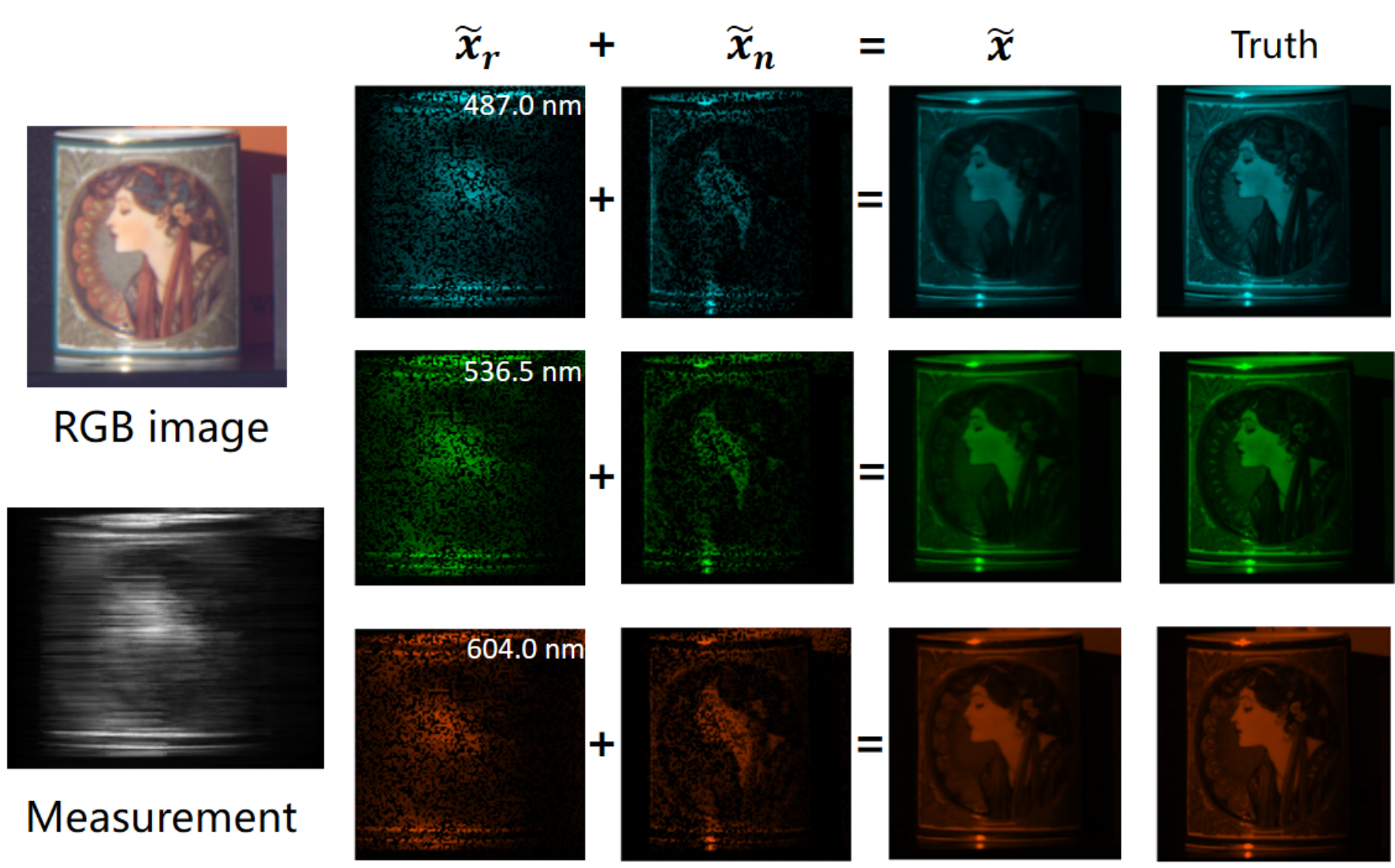}
    \caption{\textbf{Deep range-null space decomposition for hyperspectral image reconstruction}. The picture shows the main reconstruction ideas of the Range-Null Space Decomposition (RND) method. RND-SCI respectively reconstructs the data-fidelity (range-space) term $\tilde{\boldsymbol{x}}_r$ and the regularization (null-space) term $\tilde{\boldsymbol{x}}_n$. By adding the two parts, we get the reconstructed images $\tilde{\boldsymbol{x}}$.}
    \label{fig:overview}
\end{figure}

Compared to RGB images, hyperspectral images (HSIs) consist of numerous discrete bands to convey more spectral information beyond the visible. Therefore, they are a formidable tool in the realm of remote sensing~\cite{lu2020recent}, and visual recognition~\cite{kim20123d,xu2015anomaly}. However, capturing hyperspectral images at high spatial and spectral resolution at a fast speed is challenging with traditional imaging systems. To alleviate the problem, the coded aperture snapshot spectral imaging (CASSI) system~\cite{wagadarikar2008single} (Figure \ref{fig:cassi}), based on compressed sensing technology~\cite{1614066}, has been proposed and developed. The system modulates the 3D HSI signals via a coded aperture and compresses it to a 2D measurement. The main challenge in compressive spectral imaging is to design a reliable and fast algorithm to recover HSIs from a 2D measurement.

Traditional model-based methods~\cite{wang2016adaptive,kittle2010multiframe,liu2018rank,yuan2016generalized,kittle2010multiframe} attempt to construct a series of hand-crafted features and regularization by analyzing the imaging formation. However, the methods need to tune parameters manually and the generalization suffers from model capacity. One of the simple modeling approaches, an end-to-end method~\cite{Miao_2019_ICCV,meng2020end,hu2022hdnet,mst}, formulates the HSI reconstruction as a regression problem, which can achieve state-of-the-art reconstruction results thanks to the deep neural networks. The disregard for the priors and principles of CASSI systems in those methods hinders their adaptability to diverse real-world scenarios. Also, this kind of method loses flexibility and interpretability between spatial precision and speed in different scenarios.

To address the aforementioned challenges, deep unfolding methods~\cite{cai2022degradation,wang2020dnu,meng2020gap,ma2019deep} offer a promising approach to combine the principle of CASSI signal encoding to design an HSI decoder network. A common practice in these methods is to formulate the reconstruction from a bayesian perspective:
\begin{equation}
    \label{eq:optim}
    \tilde{\boldsymbol{x}}= \mathop{\arg\min}\limits_{\boldsymbol{x}}\left \|\boldsymbol{y}-\mathbf{\Phi}\boldsymbol{x}\right \|^2_2+ \lambda R(\boldsymbol{x}).
\end{equation}
The first data-fidelity term $ \left \|\boldsymbol{y}-\mathbf{\Phi}\boldsymbol{x}\right \|^2_2$ 
 optimizes data consistency, ensuring that the reconstructed results conform to the CASSI system forward process. While the second image-prior term $\lambda R(\boldsymbol{x})$ regularizes the reconstructed result with prior knowledge of HSI distribution. Compared to previous methods, they have made great progress in terms of interpretability and model flexibility. However, Eq.(\ref{eq:optim}) is a non-convex optimization problem which makes it difficult to design a convergent algorithm. Moreover, it's hard to balance data consistency and regularization for a favorable reconstruction.

In this work, we apply the Range-Null space decomposition to the HSI reconstruction. It offers a new perspective to handle data and regularization terms. The data consistency is only determined by the range-space contents and can be calculated without data-driven learning. Hence, the key problem turns to finding a proper null-space solution that makes the reconstruction performance satisfy HSI real scene distribution. Furthermore, we improve the SAUNet~\cite{wang2023simple} based on Range-Null space decomposition and achieve better reconstruction accuracy with faster reconstruction speed than other real-time HSI reconstruction systems. It can process \textbf{91} measurements per second at Tesla V100, approximately doubling the speed of CST-S~\cite{cai2022coarse}. In summary, our main contributions are \\
(1) We apply the Range-Null Space Decomposition method and design a novel framework named RND-SCI, for HSI reconstruction.\\
(2) Thanks to the flexibility of RND-SCI, different strategies like training from scratch, fine-tuning and Plug-and-Play (PnP) used RND-SCI consistently lead to improved performance in four different HSI algorithms. \\
(3) Based on the RND-SCI, we propose an extremely fast HSI reconstruction network, RND-SAUNet.\\
(4) We launch a comprehensive comparison between RND-SAUNet and other different fast HSI reconstruction methods.

\section{Related Work}

\subsection{Compressive Spectral Image Reconstruction}
Existing works tackle the ill-posed inverse problems from different perspectives. Traditional methods~\cite{kittle2010multiframe,liu2018rank,yuan2016generalized,kittle2010multiframe,tan2015compressive} rely on hand-crafted image priors, \eg total variation (TV)~\cite{yuan2016generalized}, low-rank~\cite{liu2018rank}, and sparsity~\cite{wagadarikar2008single}. However, these methods generally have limited performance and sometimes suffer from long-time consumption. 

End-to-end methods~\cite{cai2022coarse,mst,meng2020snapshot,hu2022hdnet} can leverage the advantages of neural networks to improve the representation ability. Recently, owing to convolutional neural network limits of long-range dependencies, the transformer architecture has great attention and can recover the images that match the real scenes. However, the methods have little literal theoretical proven~\cite{yuan2021snapshot} and lack the flexibility to balance between the reconstruction precision and speed in real-world usage.

Deep unfolding methods~\cite{cai2022degradation,wang2020dnu,meng2020gap,ma2019deep,huang2021deep}, different from the above methods, employ deep neural network as a denoising model in an optimization solver, such as ADMM~\cite{boyd2011distributed}, HQS~\cite{he2013half}, PGD~\cite{beck2009fast}, \etc. The methods effectively boost the flexibility and interpretability of the principle of CASSI. However, it is difficult to design a non-convex optimization framework and ensure its convergence and the interpretation of such methods is still lack.

\subsection{Range-Null Space Decomposition for inverse imaging}
Schwab \etal~\cite{schwab2019deep} proposes the method for solving inverse problems and analyzes the rate of convergence. Chen 
 \etal~\cite{chen2020deep} analytically decomposes the reconstruction result into range-space and null-space then fit them respectively. Morteza \etal~\cite{8417964} apply the theory to MRI with generative adversarial neural network (GAN) and achieve great performance. Wang \etal~\cite{wang2022gan} reveal a concise and fast super-resolution network based on the range-null space decomposition framework.

However, as for hyperspectral image reconstruction, the sensing matrix $\boldsymbol{\Phi}$ is a fat and sparse matrix, which leads to solving the pseudo-inverse difficulty. In this paper, we propose a simple and efficient approach according to the properties of the matrix, by conducting several matrix dot product operations to solve pseudo-inverse of the sensing matrix.

\section{Method}
\subsection{Preliminaries}

\begin{figure*}[t]
    \centering
    \includegraphics[width=\linewidth,scale=1.00]{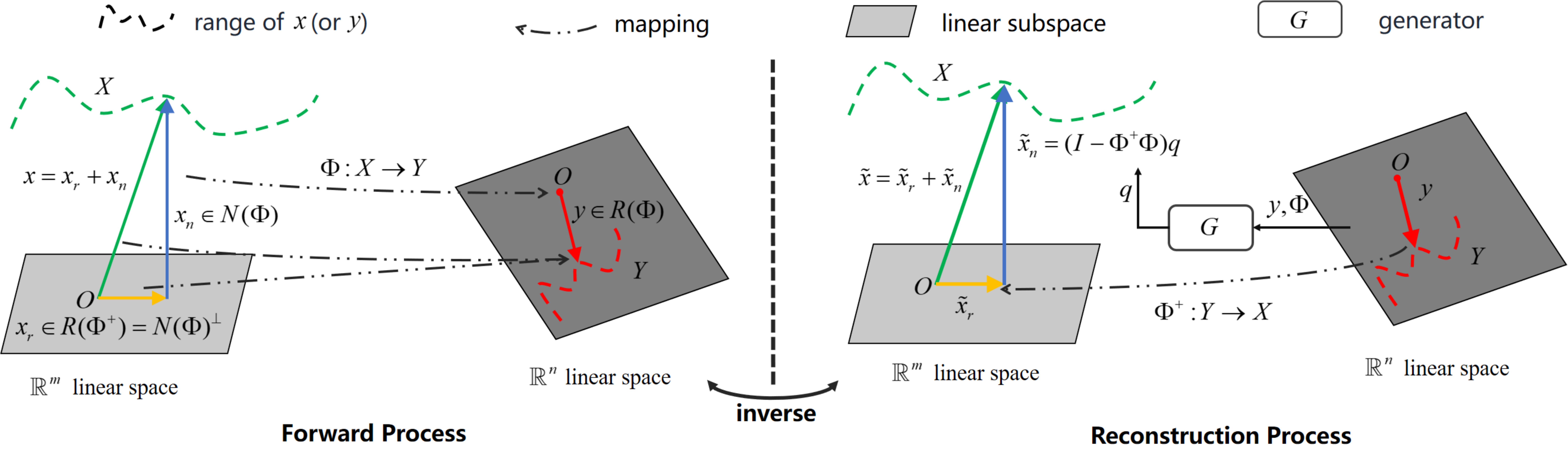}
    \caption{\textbf{The schematic diagram of Range-Null Space Decomposition.} According to the forward process $\boldsymbol{y}=\mathbf{\Phi}\boldsymbol{x}$, the $\mathbb{R}^m$ linear space can be uniquely decomposed into two orthogonal subspaces, range space and null space. During the reconstruction process, the range-space reconstructed object is constructed by the pseudo-inverse matrix $\mathbf{\Phi}^\dagger$ corresponding to $\mathbf{\Phi}$. While the other object is to first generate a $\boldsymbol{q}$ satisfying the regularization constraint through the conditional generative network and then obtained by the null space projection operator $\boldsymbol{I}-\mathbf{\Phi}^\dagger\mathbf{\Phi}$.}
    \label{fig:rnd}
\end{figure*}

\subsubsection{CASSI System}
\begin{figure}[!th]
    \centering
    \includegraphics[width=\linewidth,scale=1.00]{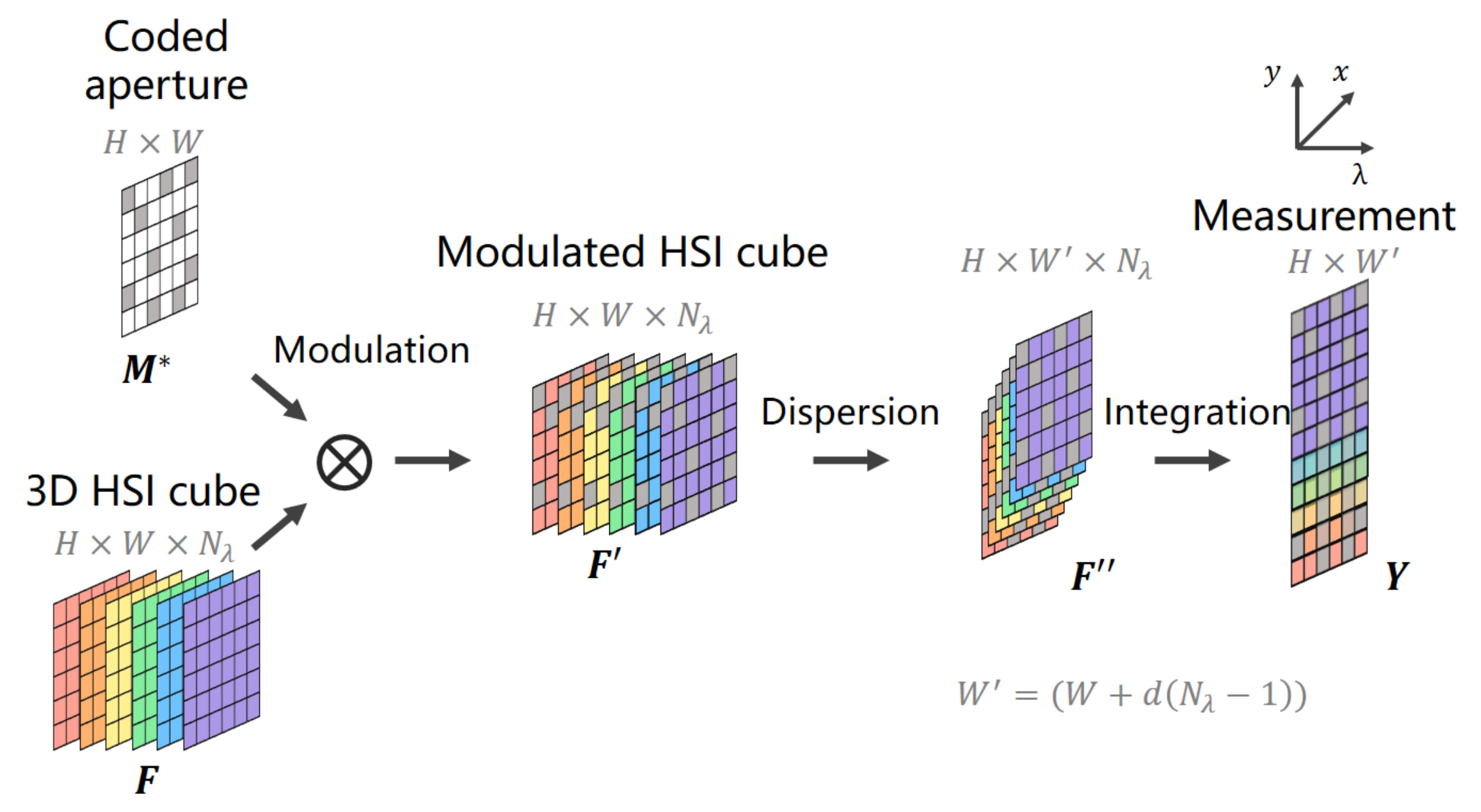}
    \caption{\textbf{Illustration of the Coded Aperture Snapshot Spectral Imaging (CASSI) system.} HSIs are handled by modulation, dispersion, and integration operations to obtain the 2D measurements.}
    \label{fig:cassi}
\end{figure}

A concise spectral image compression pipeline of the CASSI System is shown in Figure \ref{fig:cassi}. The spectral information is modulated by a pre-defined coded aperture mask $\boldsymbol{M}^*$ to modulate $\boldsymbol{F}$ and then dispersed by a dispersive prism. Finally, the signals are detected by the detector. Mathematically, we denote $\boldsymbol{M}^* \in \mathbb{R}^{H\times W}$ as coded aperture and $\boldsymbol{F}\in\mathbb{R}^{H\times W\times N_{\lambda}}$ as original 3D HSI cube. To simplify, we denote  $\boldsymbol{M}\in\mathbb{R}^{H\times(W+ d(N_{\lambda} - 1))\times N_\lambda}$ and $\boldsymbol{F''}\in\mathbb{R}^{H\times(W+ d(N_{\lambda} - 1))\times N_\lambda}$ as  3D shifted version of mask $\boldsymbol{M}^*$ and 3D HSI cube $\boldsymbol{F}$, where $d$ represents the shifting step. Their relationship is as follows:
 
\begin{equation}
    \boldsymbol{M}(u,v,n_{\lambda})=\boldsymbol{M}^*(x,y+d(\lambda_n-\lambda_c))
\end{equation}
\begin{equation}
    \boldsymbol{F''}(u,v,n_{\lambda})=\boldsymbol{F}(x,y+d(\lambda_n-\lambda_c),n_\lambda)
\end{equation}
where $(u, v)$ indicates the position on the detector plane coordinate system. $\lambda_n$ represents the wavelength of the $n_\lambda$-th spectral channel, and $\lambda_c$ is the base wavelength that does not shift after passing a dispersive prism. So the measurement $\boldsymbol{Y}$ is formulated as 
\begin{equation}
\label{eq:all_cassi}
\boldsymbol{Y}=\sum_{n_\lambda=1}^{N_\lambda}\boldsymbol{F''}(:,:,n_\lambda)\odot\boldsymbol{M}(:,:,n_{\lambda})
\end{equation}

\begin{figure}[!h]
    \centering
    \includegraphics[width=\linewidth,scale=1.00]{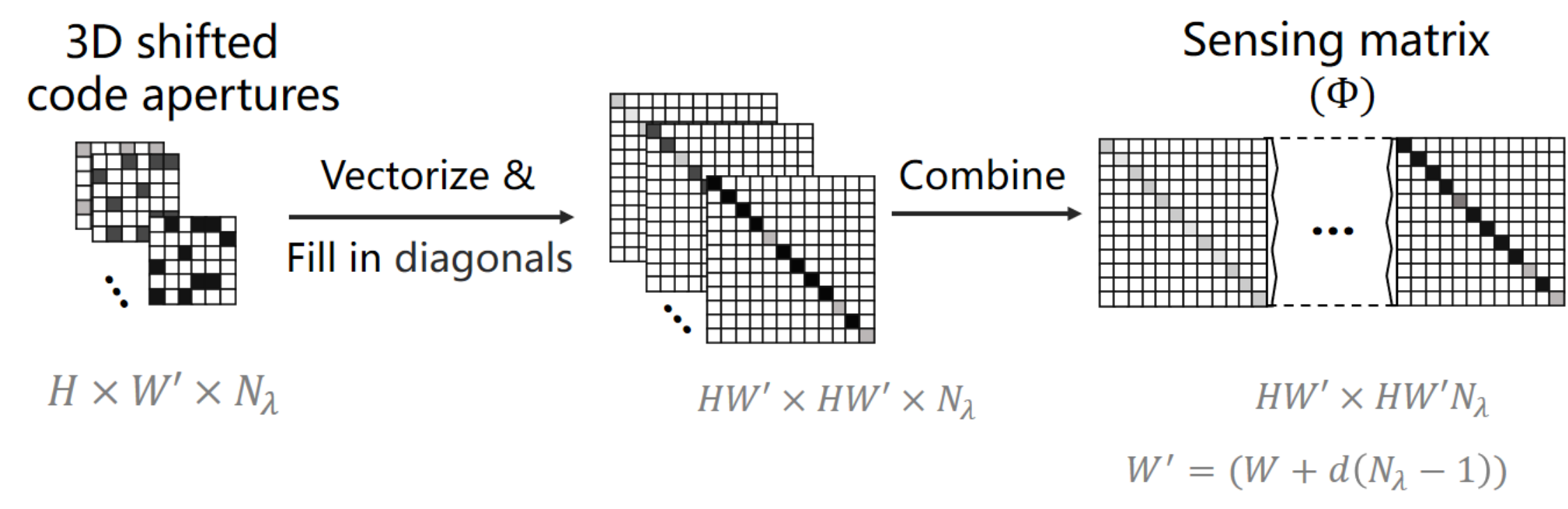}
    \caption{\textbf{The transition process from coded aperture to sensing matrix}. To generate the sensing matrix, the 3D shifted coded aperture is first vectorized and filled in each diagonal of zero matrices and then combine with the number of spectral band diagonal matrices.}
    \label{fig:Phi}
    \vspace{-5mm}
\end{figure}

\vspace{-3mm}
\paragraph{Vectorization.} Let $\mathrm{vec}(\cdot)$ represent the matrix vectorization, \ie, concatenates all the columns of a matrix as a single vector. Define $\boldsymbol{y}=\mathrm{vec}(\boldsymbol{Y})\in\mathbb{R}^n$ where  $n=H(W+d(N_\lambda-1))$. As for original 3D HSI cube, we denote the $\boldsymbol{x}_{n_\lambda}=\mathrm{vec}(\boldsymbol{F''}(:,:,n_\lambda))$. 
So the vector $\boldsymbol{x}$ is
\begin{equation}
\boldsymbol{x}=\mathrm{vec}([\boldsymbol{x}_1,\boldsymbol{x}_2,..., \boldsymbol{x}_{N_\lambda}])\in\mathbb{R}^{n{N_\lambda}}.    
\end{equation}
Similar to above steps, as shown in Figure~\ref{fig:Phi}, we denote the sensing matrix $\mathbf{\Phi}\in R^{n\times nN_\lambda}$ as
\begin{equation}
\label{eq:phi}
\mathbf{\Phi}=[\mathbf{\Phi}_1,\mathbf{\Phi}_2,...,\mathbf{\Phi}_{N_\lambda}],   
\end{equation}
where $\mathbf{\Phi}=\mathrm{diag}(\mathrm{vec}(\mathrm{\boldsymbol{M}(:,:,n_\lambda)))}$. The vectorized version of Eq.(\ref{eq:all_cassi}) is rewritten as
\begin{equation}
    \label{eq: vector}
    \boldsymbol{y}=\mathbf{\Phi}\boldsymbol{x}.
\end{equation}

\subsubsection{Range-Null Space Decomposition}

Figure~\ref{fig:rnd} shows the main idea of the Range-Null space decomposition. A linear imaging system can be described as Eq.~\ref{eq: vector}, where $\boldsymbol{y}\in\mathbb{R}^n$, $\boldsymbol{x}\in\mathbb{R}^m$ and $\mathbf{\Phi} : \mathbb{R}^m \rightarrow \mathbb{R}^n$ is a mapping between a linear spaces $\mathbb{R}^n$ and $\mathbb{R}^m$. We define the null space of the $\mathbf{\Phi}$ as $N(\mathbf{\Phi})$, which satisfies
\begin{equation}
    N(\mathbf{\Phi}) = \{\boldsymbol{x}_n|\mathbf{\Phi}\boldsymbol{x}_n = 0\}
\end{equation}

And we treat the vector $\boldsymbol{x}$ as two parts: $\boldsymbol{x}_r$ and $\boldsymbol{x}_n$, where $\boldsymbol{x}_n {\in} N(\mathbf{\Phi})$ and $\boldsymbol{x}_r {\in} N(\mathbf{\Phi})^{\perp}$. And $N(\cdot)^{\perp}$ indicates the subspace perpendicular to $N(\cdot)$, i.e. the range subspace. The projection operation maps the two vectors to $\boldsymbol{y}$ and zero vector in $\mathbb{R}^m$ space, and the sum of the mapping vectors exactly equals $\boldsymbol{y}$.

During the reconstruction process, we note the pseudo-inverse of $\mathbf{\Phi}$ as $\mathbf{\Phi}^\dagger$. We can prove that the general solution of the Eq.~\ref{eq: vector} is
\begin{equation}
    \tilde{\boldsymbol{x}} = \mathbf{\Phi}^\dagger\boldsymbol{y} +(\boldsymbol{I}- \mathbf{\Phi}^\dagger\mathbf{\Phi})\boldsymbol{q},
    \label{eq:sol}
\end{equation}
where $\boldsymbol{I}$ is a unit matrix and $\boldsymbol{q}$ is a random vector in $\mathbb{R}^m$. Since $\mathbf{\Phi}\mathbf{\Phi}^\dagger\mathbf{\Phi}\equiv\mathbf{\Phi}$, we consider $\mathbf{\Phi}\mathbf{\Phi}^\dagger\mathbf{\Phi}$ as the project operator that maps samples into the range space of $\mathbf{\Phi}$. Furthermore, $\mathbf{\Phi}(\boldsymbol{I} - \mathbf{\Phi}^\dagger\mathbf{\Phi})\equiv\boldsymbol{0}$ so that $(\boldsymbol{I}- \mathbf{\Phi}^\dagger\mathbf{\Phi})$ can be seen as the operator that maps samples into null-space of $\mathbf{\Phi}$. 

Based on above, we use $\tilde{\boldsymbol{x}}_r=\mathbf{\Phi}^\dagger\boldsymbol{y}$ to represent the reconstructed range-space results while $\tilde{\boldsymbol{x}}_n=(\boldsymbol{I}- \mathbf{\Phi}^\dagger\mathbf{\Phi})\boldsymbol{q}$ to represent the null-space results. We can employ a conditional generative network to produce an approximate and satisfactory $\boldsymbol{q}$ via $(Y,\mathbf{\Phi})$. Because $\tilde{\boldsymbol{x}_r}$ is determined, this pipeline is essential to design a conditional generative network to learn a mapping from $(Y,\mathbf{\Phi})$ to the null space of $X$.

\subsection{Overall Architecture}
\begin{figure}[t]
    \centering
    \includegraphics[width=\linewidth,scale=1.00]{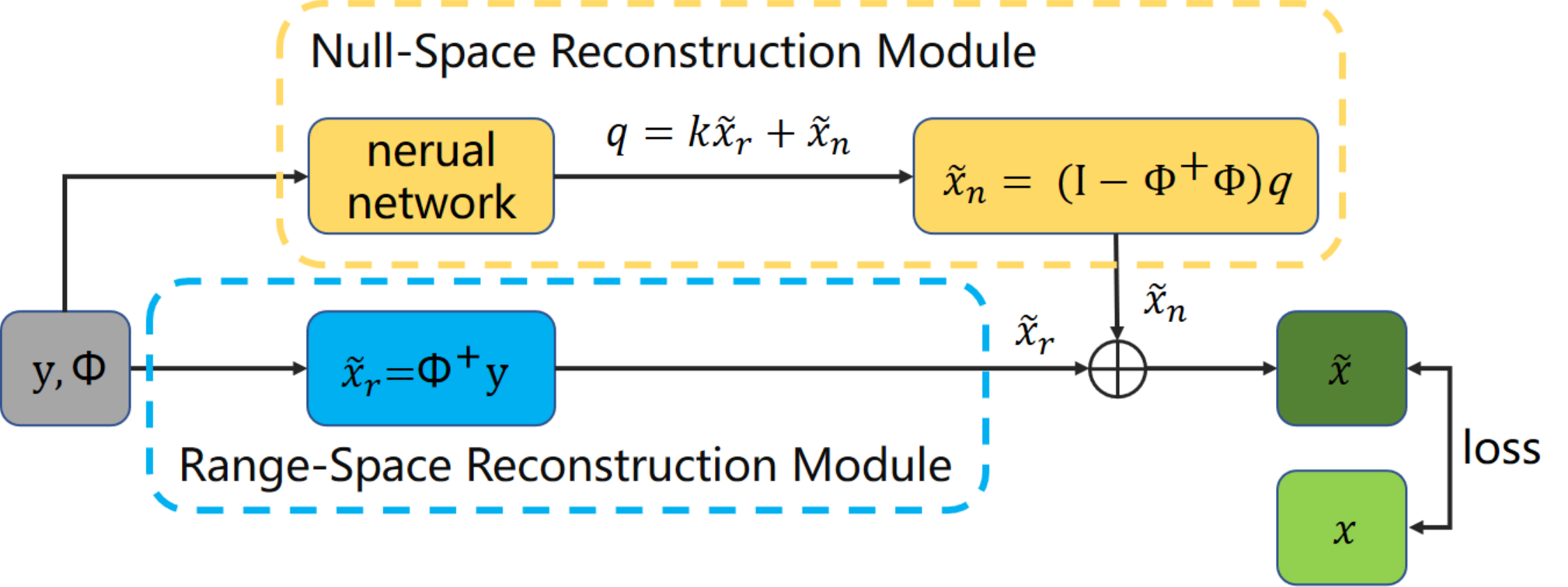}
    \caption{\textbf{RND-SCI Framework}. The RND-SCI Framework consists of two modules, the range-space reconstruction module(the blue part) and the null-space reconstruction module. The two modules respectively reconstruct the range-space target and null-space target. Finally, the solution target is obtained by adding the reconstruction contents of modules.}
    \label{fig:framework}
\end{figure}

Our framework is shown as Figure~\ref{fig:framework}. Based on the theory of range-null space decomposition, we design the range-space reconstruction module and the null-space reconstruction module to produce the range-space value $\tilde{\boldsymbol{x}}_r$ and null-space value $\tilde{\boldsymbol{x}}_n$ respectively. The final output $\tilde{\boldsymbol{x}}$ is the dot sum of the two reconstructed images. 

The range-space reconstruction module consists of a matrix linear operation $\mathbf{\Phi}^{\dagger} \boldsymbol{y}$, where $\mathbf{\Phi}^\dagger$ is the pseudo-inverse of the sensing matrix $\mathbf{\Phi}$ (in Figure \ref{fig:Phi}). The purpose of this module is to guarantee that the reconstruction results are consistent with the degradation process, i.e. $\tilde{\boldsymbol{x}}_r$ is one of the solutions of Eq.~\ref{eq: vector}. While the null-space reconstruction module needs a conditional generative neural network to produce an eligible vector $\boldsymbol{q}$ according to the HSI prior constraints. In order to make the final results meet the requirements of data consistency and regularization, the vector $\boldsymbol{q}$ is left multiplied by a null space projection matrix to project $\boldsymbol{q}$ into the null space of $\mathbf{\Phi}$ and acquire $\tilde{\boldsymbol{x}}_n$. According to the properties of $\boldsymbol{I} - \mathbf{\Phi}^\dagger\mathbf{\Phi}$, the reconstruction target of the neural network should be 
\begin{equation}
    \boldsymbol{q} = k \boldsymbol{x}_r + \boldsymbol{x}_n,
    \label{eq:q}
\end{equation}
where $k$ is an arbitrary real number. When $k$ is set to 1, the output of the conditional generative neural network is $\tilde{\boldsymbol{x}}$. Therefore, each existing compressive spectral reconstruction network which is viewed as a conditional generative neural network can insert the RND framework.

So the key problem is \textbf{how to solve the pseudo-inverse of $\boldsymbol{\Phi}$ to solve the range-space term.} It is common to use singular value decomposition(SVD) \cite{klema1980singular} to compute the analytical solution, that is 
\begin{equation}
    \boldsymbol{\Phi} = \mathbf{U}\mathbf{\Sigma}\mathbf{V}^\top, \quad \boldsymbol{\Phi}^\dagger = \mathbf{V}\mathbf{\Sigma}^{-1}\mathbf{U}^\top,
\end{equation}
where $\mathbf{U}$ and $\mathbf{V}$ are orthogonal matrices and $\mathbf{\Sigma}$ is a diagonal matrix with eigenvalues as its diagonal elements. However, it is hard to compute directly on account of the sparse and fat $\boldsymbol{\Phi}$. For example, $H$ = 256, $W$ = 256, $N_\lambda$ = 28 and store the matrix $\boldsymbol{\Phi}$ as 32 bit float-point number, it takes about \textbf{656 GB} memory storage. However, under the same conditions, the 3D-shifted coded aperture (left in Figure~\ref{fig:Phi})  takes only about \textbf{8.4 MB} memory storage. A natural idea is to simplify this pseudo-inverse with the help of the CASSI principle.

As shown in Figure~\ref{fig:Phi}, it is worth noting that the special structure of the sensing matrix $\boldsymbol{\Phi}$ will lead to $\boldsymbol{\Phi}\boldsymbol{\Phi}^\top$ is a diagonal matrix and the each of the diagonal elements is the sum of the squares of the 3D shifted coded aperture along the spectral band dimension, that is
\begin{equation}
    \boldsymbol{\Phi}\boldsymbol{\Phi}^\top(u + vH, u + vH) = \sum^{N_\lambda}_{n_\lambda=1} \boldsymbol{M}^2(u,v,n_\lambda).
\end{equation}

In engineering, it's easy to guarantee matrix $\boldsymbol{\Phi}\boldsymbol{\Phi}^\top$ is a full rank matrix, that is  $rank(\boldsymbol{\Phi}\boldsymbol{\Phi}^\top)=HW$. Due to $rank(\boldsymbol{\Phi}\boldsymbol{\Phi}^\top)=rank(\boldsymbol{\Phi}^\top\boldsymbol{\Phi})=rank(\boldsymbol{\Phi})$, it is easy to prove that matrix $\boldsymbol{\Phi}$ is a row full rank matrix. A simple way to solve the pseudo-inverse of a matrix $\boldsymbol{\Phi}$ is
\begin{equation}
    \mathbf{\Phi}^\dagger = \mathbf{\Phi}^\top(\mathbf{\Phi}\mathbf{\Phi}^\top)^{-1}.
\end{equation}
Note the $(\mathbf{\Phi}\mathbf{\Phi}^\top)^{-1} = \mathbf{\Sigma}^{-1}$. So we simplify the Eq.~\ref{eq:sol} as follows:
\begin{equation}
    \tilde{\boldsymbol{x}} = \mathbf{\Phi}^\top(\mathbf{\Sigma}^{-1}\boldsymbol{y}) + [\boldsymbol{q} - \mathbf{\Phi}^\top\mathbf{\Sigma}^{-1}(\mathbf{\Phi}\boldsymbol{q})].
    \label{eq:sol}
\end{equation}

Notice that we just need a couple of element-wise computations to solve for $\boldsymbol{x}$ from $\boldsymbol{q}$, $\boldsymbol{\Phi}$ and $\boldsymbol{y}$. Hence, the method for HSI reconstruction is very efficient and accurate when the vector $\boldsymbol{q}$ is reconstructed exactly.

\subsection{Fast HSI Reconstruction}
In order to maximize the performance and speed up the reconstruction process, we design RND-SAUNet, combining with improved SAUNet~\cite{wang2023simple} and RND. 

SAUNet unfolds the ADMM optimization algorithm and incorporates a set of learnable parameters for relaxing the residual terms. Each sub-network comprises a linear layer and a U-shaped denoiser with Transformer-style CNN~\cite{hou2022conv2former}. Although it can be quite fast and achieve high accuracy performance. However, We are able to make some following adjustments to boost the speed and stability of the network.
\paragraph{Better initialization.} 
As for the original SAUNet, we first repeat $\boldsymbol{y}$ in spectral dimension. Note the repeated tensor as $Z$. The next 3D shift operation can be formulated as 
\begin{equation}
Z_{s}(u,v,c)  = 
\begin{cases}
Z(u,v-dc,c), &0 \le v-dc\le W,\\
0, & v - dc \le 0,
\end{cases}
\end{equation}
where $Z_{s}$ is the tensor $Z$ after the 3D shift operation and c is the index of spectral dimension.

But the shifted operation with zero padding may lack some spectral characteristic information, which leads to generating a poor initialization $z^0$. Hence, the poor initialization further increases the burden of the subsequent denoising network.

\begin{figure}[!h]
    \centering
    \includegraphics[width=\linewidth,scale=1.00]{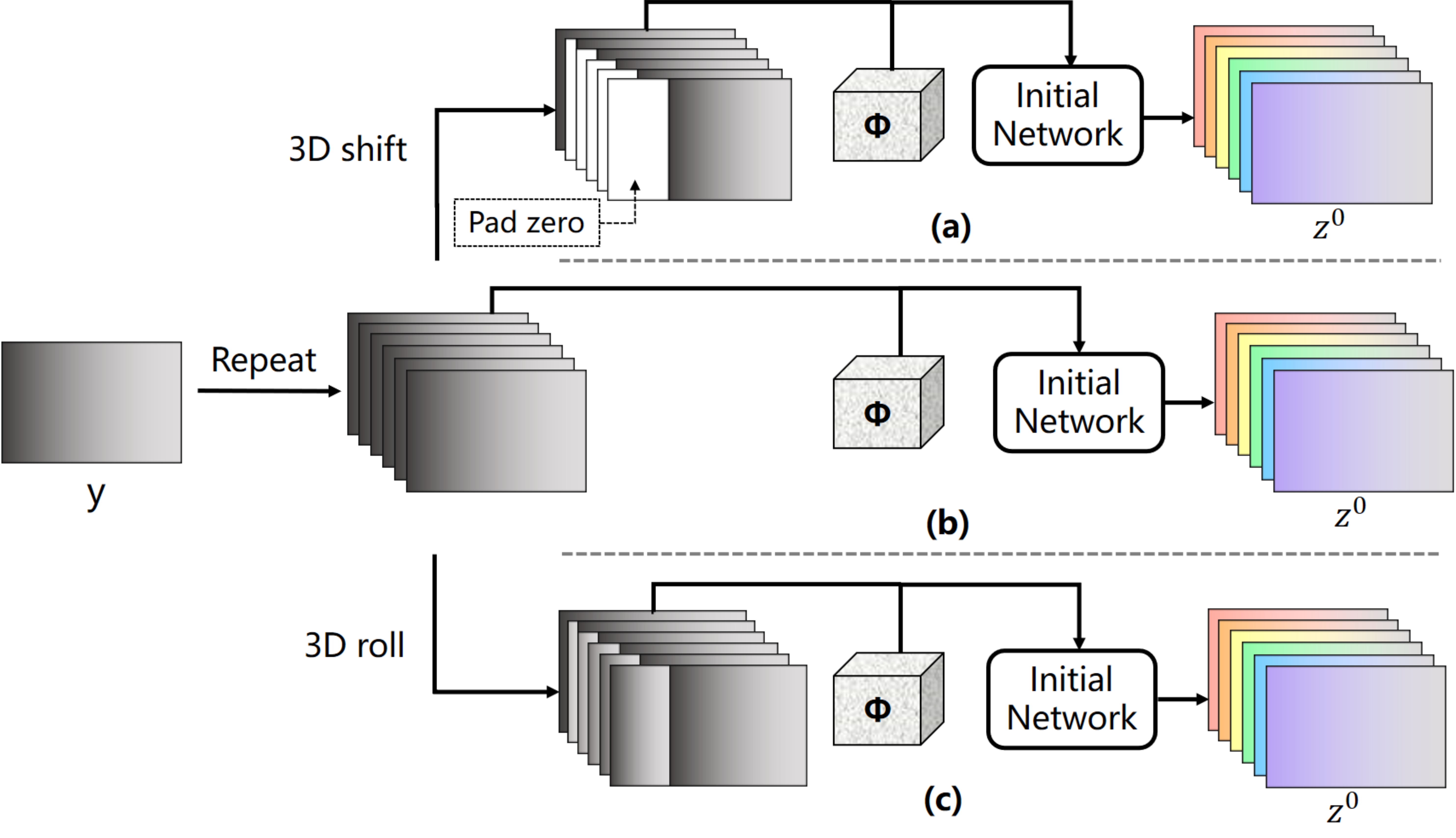}
    \caption{\textbf{Three kinds of initial network design}. (a) for original SAUNet, (b) and (c) are more stable initial network designs. (b) is that direct using repeated y as initial network input, while (a) and (c) generate input via the two different operations, 3D shift, and 3D roll.}
    \label{fig:init}
\end{figure}

One improvement (Figure~\ref{fig:init}) is to use $Z$ as input to produce $z^0$. However, the matrix $\Phi$ is shifted based on the following operations. This may lead to confusing interactions between $\Phi$ and affect the performance of the initialization $z^0$.

\begin{figure}[!h]
    \centering
    \includegraphics[width=\linewidth,scale=1.00]{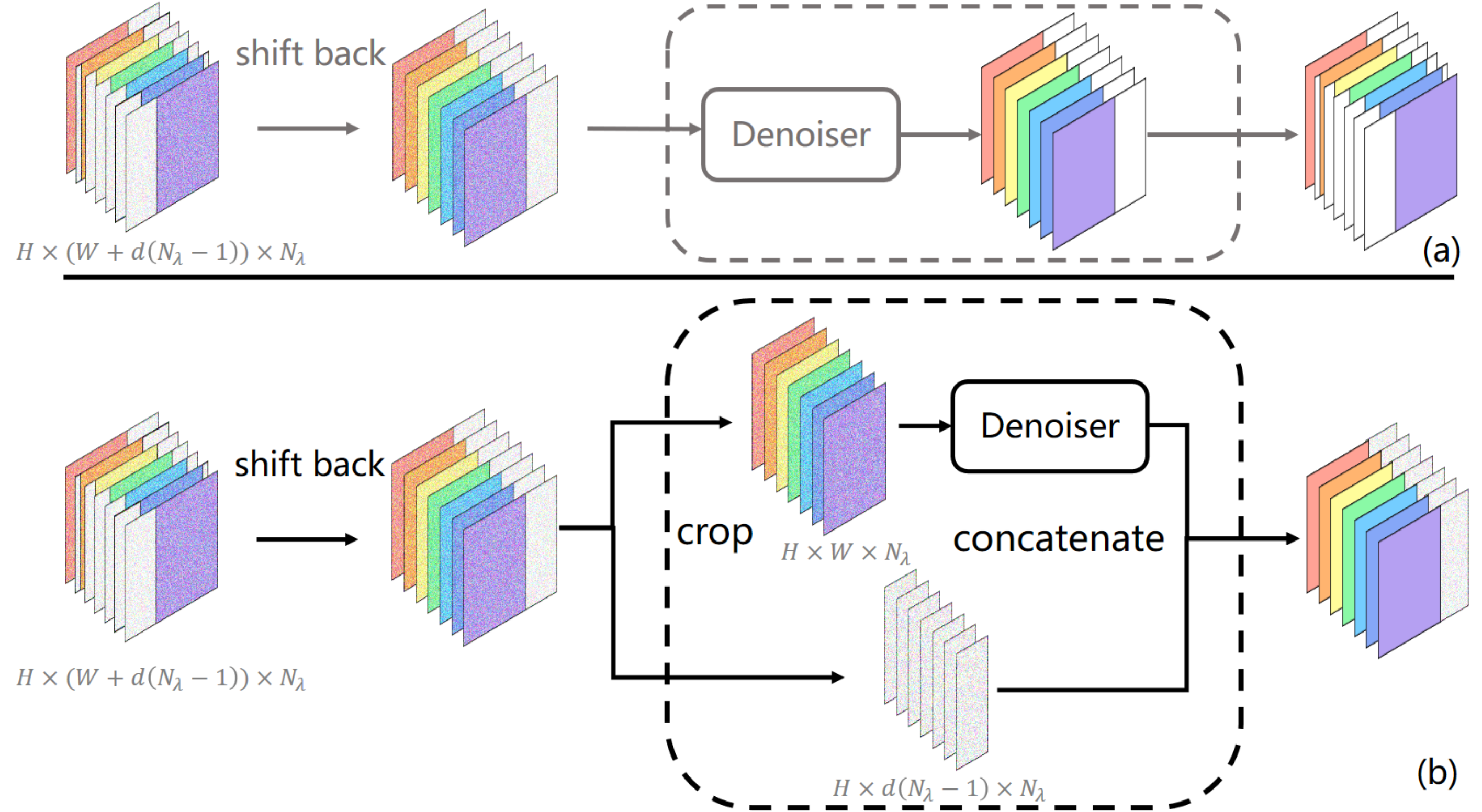}
    \caption{\textbf{Crop the denoiser input}.}
    \label{fig:denoiser}
    \vspace{-4mm}
\end{figure}

So we finally come up with a new method to input approach to generate initialization. As shown in (c) of Figure~\ref{fig:init}, after the 3D roll for the tensor $Z$, the rolled tensor $Z_r$ can be represented as 
\begin{equation}
    Z_r(u,v,c) = Z(u, (v-dc+W)\% W,c),
\end{equation}
where the operation $\%$ is a modulo operation.
\vspace{-3mm}
\paragraph{Cropped input into denoiser.}
For a fast reconstruction, as shown in Figure~\ref{fig:denoiser}, we depart from the original denoiser input as HSIs with noise (the colored part) and the shifted part from HSIs (the white part). One intuitive idea is to feed noisy HSI signals into the network. This method not only can speed up the network and reduce the computational complexity but also does not cause a significant reduction in accuracy.

\begin{table*}[!h]
    \renewcommand{\arraystretch}{1.0}
    \newcommand{\tabincell}[2]{\begin{tabular}{@{}#1@{}}#2\end{tabular}}
    \centering
    \resizebox{0.9\textwidth}{!}
    {
    \begin{tabular}{cccccccccccccc}
       \toprule
       \rowcolor{lightgray}
       Algorithms & Params & GFLOPs & S1 & S2 & S3 & S4 & S5 & S6 & S7 & S8 & S9 & S10 & Avg \\
       \midrule
        DGSMP~\cite{huang2021deep}
        & 3.76M
        & 646.65
        & \tabincell{c}{33.26\\0.915}
        & \tabincell{c}{32.09\\0.898}
        & \tabincell{c}{33.06\\0.925}
        & \tabincell{c}{40.54\\0.964}
        & \tabincell{c}{28.86\\0.882}
        & \tabincell{c}{33.08\\0.937}
        & \tabincell{c}{30.74\\0.886}
        & \tabincell{c}{31.55\\0.923}
        & \tabincell{c}{31.66\\0.911}
        & \tabincell{c}{31.44\\0.925}
        & \tabincell{c}{32.63\\0.917}
        \\
        \midrule
        TSA-Net~\cite{meng2020end}
        & 44.25M
        & 110.06
        & \tabincell{c}{33.48\\0.919}
        & \tabincell{c}{32.70\\0.900}
        & \tabincell{c}{34.63\\0.942}
        & \tabincell{c}{41.26\\0.973}
        & \tabincell{c}{30.90\\0.921}
        & \tabincell{c}{32.33\\0.934}
        & \tabincell{c}{31.42\\0.902}
        & \tabincell{c}{30.62\\0.924}
        & \tabincell{c}{32.65\\0.926}
        & \tabincell{c}{29.90\\0.896}
        & \tabincell{c}{32.99\\0.924}
        \\
        \midrule
        GAP-Net~\cite{meng2020gap}
        & 4.27M
        & 78.58
        & \tabincell{c}{33.74\\0.911}
        & \tabincell{c}{33.26\\0.900}
        & \tabincell{c}{34.28\\0.929}
        & \tabincell{c}{41.03\\0.967}
        & \tabincell{c}{31.44\\0.919}
        & \tabincell{c}{32.40\\0.925}
        & \tabincell{c}{32.27\\0.902}
        & \tabincell{c}{30.46\\0.905}
        & \tabincell{c}{33.51\\0.915}
        & \tabincell{c}{30.24\\0.895}
        & \tabincell{c}{33.26\\0.917}
        \\
        \midrule
        ADMM-Net~\cite{ma2019deep}
        & 4.27M
        & 78.58
        & \tabincell{c}{34.12\\0.918}
        & \tabincell{c}{33.62\\0.902}
        & \tabincell{c}{35.04\\0.931}
        & \tabincell{c}{41.15\\0.966}
        & \tabincell{c}{31.82\\0.922}
        & \tabincell{c}{32.54\\0.924}
        & \tabincell{c}{32.42\\0.896}
        & \tabincell{c}{30.74\\0.907}
        & \tabincell{c}{33.75\\0.915}
        & \tabincell{c}{30.68\\0.895}
        & \tabincell{c}{33.58\\0.918}
        \\
        \midrule
        MST-S~\cite{mst}
        & 0.93M
        & 12.96
        & \tabincell{c}{34.71\\0.930}
        & \tabincell{c}{34.45\\0.925}
        & \tabincell{c}{35.32\\0.943}
        & \tabincell{c}{41.50\\0.967}
        & \tabincell{c}{31.90\\0.933}
        & \tabincell{c}{33.85\\0.943}
        & \tabincell{c}{32.69\\0.911}
        & \tabincell{c}{31.69\\0.933}
        & \tabincell{c}{34.67\\0.939}
        & \tabincell{c}{31.82\\0.926}
        & \tabincell{c}{34.26\\0.935}
        \\
        \midrule
        CST-S~\cite{cai2022coarse}
        & 1.20M
        & 11.67
        & \tabincell{c}{34.78\\0.930}
        & \tabincell{c}{34.81\\0.931}
        & \tabincell{c}{35.42\\0.944}
        & \tabincell{c}{41.84\\0.967}
        & \tabincell{c}{32.29\\0.939}
        & \tabincell{c}{34.49\\0.949}
        & \tabincell{c}{33.47\\0.922}
        & \tabincell{c}{\textbf{32.89}\\0.945}
        & \tabincell{c}{34.96\\0.944}
        & \tabincell{c}{32.14\\0.932}
        & \tabincell{c}{34.71\\0.940}
        \\
        \midrule
        HDNet~\cite{hu2022hdnet}
        & 2.37M
        & 157.76
        & \tabincell{c}{35.14\\0.935}
        & \tabincell{c}{\textbf{35.67}\\\textbf{0.940}}
        & \tabincell{c}{36.03\\0.943}
        & \tabincell{c}{42.30\\0.969}
        & \tabincell{c}{\textbf{32.69}\\\textbf{0.946}}
        & \tabincell{c}{\textbf{34.46}\\\textbf{0.952}}
        & \tabincell{c}{\textbf{33.67}\\0.926}
        & \tabincell{c}{32.48\\0.941}
        & \tabincell{c}{34.89\\0.942}
        & \tabincell{c}{\textbf{32.38}\\\textbf{0.937}}
        & \tabincell{c}{34.97\\0.943}
        \\
        \midrule
        SAUNet-1stg~\cite{wang2023simple}
        & 0.78M
        & 9.52
        & \tabincell{c}{34.66\\0.936}
        & \tabincell{c}{34.78\\0.933}
        & \tabincell{c}{36.74\\\textbf{0.955}}
        & \tabincell{c}{\textbf{43.33}\\\textbf{0.979}}
        & \tabincell{c}{32.14\\0.941}
        & \tabincell{c}{34.28\\\textbf{0.952}}
        & \tabincell{c}{33.29\\\textbf{0.927}}
        & \tabincell{c}{32.18\\\textbf{0.947}}
        & \tabincell{c}{35.24\\\textbf{0.950}}
        & \tabincell{c}{31.79\\0.936}
        & \tabincell{c}{34.84\\\textbf{0.946}}
        \\
        \midrule
        \rowcolor{Highlight}
        RND-SAUNet(Ours)
        & \textbf{0.78M}
        & \textbf{7.84}
        & \tabincell{c}{\textbf{34.91}\\\textbf{0.937}}
        & \tabincell{c}{35.19\\0.933}
        & \tabincell{c}{\textbf{37.12}\\\textbf{0.955}}
        & \tabincell{c}{42.28\\0.971}
        & \tabincell{c}{32.34\\0.941}
        & \tabincell{c}{34.44\\\textbf{0.952}}
        & \tabincell{c}{33.43\\0.926}
        & \tabincell{c}{32.66\\0.944}
        & \tabincell{c}{\textbf{35.56}\\0.948}
        & \tabincell{c}{32.07\\0.936}
        & \tabincell{c}{\textbf{35.00}\\0.944}
        \\
       \bottomrule
    \end{tabular}
    }
    \caption{\textbf{Comparisons between RND-SAUNet and SOTA low time-consuming methods on 10 simulation scenes (S1$\sim$S10).} Params, FLOPS, PSNR (upper entry in each cell), and SSIM (lower entry in each cell) are reported. }
    \label{tab:simu}
\end{table*}

\section{Experiments}
\subsection{Experiments Setting}
We follow TSA-Net~\cite{meng2020end} to adopt 28 wavelengths from 450nm to 650nm derived by spectral interpolation for simulation HSI reconstruction experiments.
\vspace{-3mm}
\paragraph{Datasets}
In our experiments, we choose two publicly available datasets, namely CAVE~\cite{yasuma2010generalized} and KAIST~\cite{DeepCASSI:SIGA:2017}, and to ensure a fair comparison, we utilize a mask of size $256\times256$, identical to that of TSA-Net. And the test dataset consists of 10 scenes from KAIST, while the others are used for training.
\vspace{-3mm}
\paragraph{Implementation}
We implement our model in PyTorch. All models are trained with Adam optimizer~\cite{kingma2014adam} ($\beta_1=0.9$ and $\beta_2=0.999$) for 300 epochs. The learning rate
is set to $4\times10^{-4}$ at the beginning and is halved every 50 epochs during the training procedure. We randomly crop patches with the spatial size of $256\times256$ and $660\times660$ from 3D HSI cubes as training samples for simulation and real experiments. Following TSA-Net~\cite{meng2020end}, the shifting step $d$ in dispersion is set to 2 and the batch size is 5. We set reconstructed channel $C=N_\lambda=28$. Data augmentation is made of random flipping and rotation. The training cost function is to follow the design of the original model.

Finally, we adopt PSNR and SSIM for quantitative comparison as reference image quality assessment metrics.

\begin{table*}[!h]
    \centering
    \resizebox{1.0\linewidth}{!}{
        \begin{tabular}{c|ccccccccc}
           Model
           & TSA-Net
           & DGSMP
           & GAP-Net
           & ADMM-Net
           & MST-S
           & CST-S
           & HDNet
           & SAUNet-1stg
           & \cellcolor{Highlight}RND-SAUNet(Ours)\\ \shline
            Training Hours
            & 28.5
            & 148.4
            & 48.8
            & 48.8
            & 48.6
            & 17.5
            & 34.7
            & 14.3
            & \baseline{\textbf{9.6}}
            \\
            Inference FPS
            & 32.8
            & 8.0
            & 19.0
            & 18.6
            & 23.2
            & 45.3
            & 44.8
            & 14.3
            & \baseline{\textbf{91.0}}
        \end{tabular}
    }
    \caption{\textbf{Comparisons between RND-SAUNet and 8 SOTA low time-consuming methods with their variants in training GPU hours and inference speed at a Tesla v100 GPU.} We record the forward and backward time of each model training in a GPU for 300 epochs at batch-size=5 as training hours.}
    \label{tab:speed}
\end{table*}

\subsection{Reconstruction Results}
\begin{figure*}[!h]
    \centering
    \includegraphics[width=\linewidth,scale=1.00]{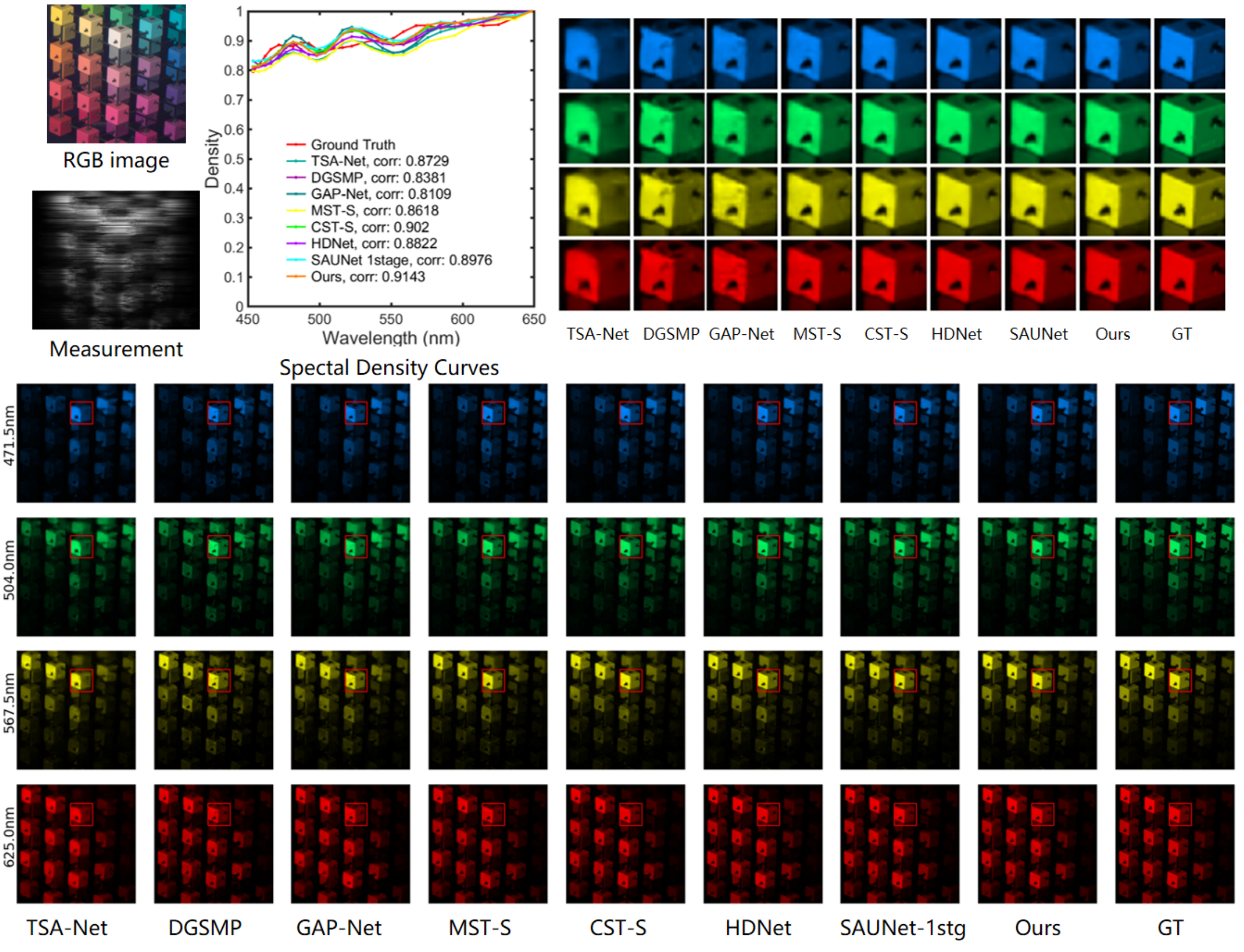}
    \caption{\textbf{Reconstructed simulation HSI comparisons of Scene 2 with 4 out of 28 spectral channels.}The top-left shows the target scene RGB image and 2D measurement. The top-middle shows the spectral
curves corresponding to the red box of the RGB image. The top-right depicts the enlarged patches corresponding to the red boxes in the
bottom HSIs. Zoom in for a better view.}
    \label{fig:simu}
\end{figure*}

Table \ref{tab:simu} compares the Params, FLOPs, PSNR, and SSIM of RND-SAUNet with other 8 SOTA low time-consuming methods. Our method RND-SAUNet surpasses other methods in both GFLOPs, params, and average PSNR/SSIM. And Table \ref{tab:speed} compares training time and test inference FPS of RND-SAUNet with other 8 SOTA low time-consuming methods in a Tesla V100 GPU. We re-train all models for 300 epochs with the toolbox of MST~\cite{mst} at 5 batch size. For more efficient computation, we use CUDA programming to optimize the 3D shift operations. RND-SAUNet sets new state-of-the-art inference and training speed scores on the simulation datasets. Figure \ref{fig:simu} depicts the simulation HSI reconstruction comparisons between our RND-SAUNet and other SOTA methods on Scene 2 with 4 (out of 28) spectral channels. RND-SAUNet has more similar spectral density curves compared to other counterparts.

\subsection{Ablation Study for RND-SAUNet}
We conduct a break-down ablation experiment to investigate the effect of each component towards higher performance or a fast speed. The results are listed in Table \ref{tab:ablation:breakdown} and Table \ref{tab:ablation:init}.
\begin{table}[!h]
\centering
\smalltablestyle{5pt}{1.05}
\begin{tabular}{ccccccc}
    crop &  init  &  RND   & PSNR & SSIM  & params (M) & GFLOPs \\
    \shline
          &       &        & 34.84 & 0.946 & 0.78M & 9.52 \\
    \checkmark& &          & 34.63 & 0.943 & 0.78M & 7.83 \\
              &\checkmark& & 34.86 & 0.946 & 0.78M & 9.52 \\
    \checkmark&\checkmark& & 34.77 & 0.946 & 0.78M & 7.84 \\
    \checkmark& &\checkmark& 34.75 & 0.942 & 0.78M & 7.83 \\
     &\checkmark&\checkmark& 35.08 & 0.945 & 0.78M & 9.52 \\ \shline
    \checkmark &\checkmark &\checkmark & \baseline{\textbf{35.00}} & \baseline{\textbf{0.944}} & \baseline{\textbf{0.78M}} & 7.84
\end{tabular}
\caption{\textbf{Ablation study} of crop, initialization and range-null decomposition.}
\label{tab:ablation:breakdown}
\end{table}
\vspace{-3mm}
\paragraph{\textbf{Influence of good initialization.}} We set three baselines, the original SAUNet-1stg, the one with cropping operation, and the RND-SAUNet without initialization improvement in Table \ref{tab:ablation:breakdown}. We can find the results increase by 0.02dB, 0.14dB, and 0.25 dB. It demonstrates that a good initialization is conducive to more accurate reconstruction performance. 
In order to further explore the impact of different initialization methods on the reconstruction results, we employ the different initial network inputs in SAUNet  with cropping denoiser input and RND-SAUNet. From Table \ref{tab:ablation:init}, we can observe approach c gets the best PSNR and SSIM in two different situations. This verifies that the 3D roll operation is more efficient for generating the  initial HSIs that retain more valid information.
\vspace{-3mm}
\paragraph{\textbf{Influence of cropping operation.}} The baselines are the original SAUNet-1stg, the one with the initial approach c in Figure \ref{fig:init}, and the one with all improvements except cropping. When we crop each denoiser input, although the PSNR metrics are lost by 0.21, 0.09 dB, and 0.08 dB, the computation overheads both decrease by 1.69 GFLOPs. This shows that crop operation can save some calculations, however, when high reconstruction accuracy is required, the information shifted from HSI is the same important.

\paragraph{\textbf{Influence of RND-SCI Framework.}} In Table, we employ the RND-SCI framework into SAUNet with cropping denoiser input, SAUNet at an initial approach c, and the one with both above. The performances are better than 0.12dB, 0.22dB, and 0.23 dB, and without much increase in computation. It illustrates the effectiveness of the RND-SCI framework SAUNet.
\begin{table}[!h]
\centering
\smalltablestyle{10pt}{1.05}
\begin{tabular}{cccc}
     init & RND & PSNR & SSIM \\
    \shline
    a &  & 34.63 &0.943 \\
    b &  & 34.76 & 0.944 \\
    c &  & 34.77 & 0.946 \\ \shline
    a &\checkmark & 34.75& 0.942 \\ 
    b &\checkmark & 34.87&  0.942\\
    c &\checkmark & \baseline{\textbf{35.00}}& \baseline{\textbf{0.944}}\\
\end{tabular}
\caption{\textbf{Ablation study} of different initialization strategies for RND-SAUNet and SAUNet with cropping denoiser input.}

\label{tab:ablation:init}
\end{table}

\begin{table}[!h]
\centering
\smalltablestyle{3.5pt}{1.05}
\begin{tabular}{l* cc|cc|cc}
    & \multicolumn{2}{c}{\footnotesize original} & \multicolumn{2}{c}{\footnotesize mask1} & \multicolumn{2}{c}{\footnotesize mask2} \\
    method & PSNR & SSIM & PSNR & SSIM & PSNR & SSIM \\
    \shline
    TSA-Net~\cite{meng2020end} & 32.99 & 0.923 & 32.68 & 0.921 & 32.91 & 0.922 \\
    MST-S~\cite{mst}           & 34.26 & 0.935 & 34.30 & 0.938 & 34.43 & 0.939 \\
    CST-S~\cite{cai2022coarse} & 34.71 & 0.940 & 34.63 & 0.945 & 34.67 & 0.944 \\  \shline
    RND-MST-S        & 34.88 & 0.942 & 34.66 & 0.940 & 34.83 & 0.942 \\
    RND-CST-S        & 34.86 & 0.943 & 34.72 & 0.941 & 34.79 & 0.942 \\ 
    RND-SAUNet       & 35.00 & 0.944 & 34.82 & 0.944 & 34.92 & 0.943 \\
\end{tabular}
\caption{\textbf{Robustness test} with different modulated masks.}
\label{tab:ablation:modulate}
\end{table}

\subsection{Generalization for Modulated Mask} In this part, we want to explore the different models for the random modulated mask. We change the mask by randomly cropping it with size $256\times256$ from the real mask of size $660\times660$ to evaluate the flexibility of the RND-SCI framework and RND-SAUNet for different signal modulations. The results are reported in Table \ref{tab:ablation:modulate}. Compared with the other models, the model with RND-SCI can achieve more accurate reconstruction performance. These results suggest that the RND-SCI framework is more robust and flexible to improve the model's HSI reconstruction.
\begin{table}[!h]
\centering
\smalltablestyle{10pt}{1.05}
\begin{tabular}{l* ccc}
     method & RND & PSNR & SSIM \\
    \shline
    MST-S &            & 34.26 & 0.940 \\
    MST-S & \checkmark & \baseline{\textbf{34.88}} & \baseline{\textbf{0.942}} \\
    \hline
    MST-M &            & 34.94 & 0.943 \\
    MST-M & \checkmark & \baseline{\textbf{35.36}} & \baseline{\textbf{0.949}} \\
    \hline
    CST-S &            & 34.71 & 0.940 \\
    CST-S & \checkmark & \baseline{\textbf{34.86}} & \baseline{\textbf{0.943}} \\
    \hline
    CST-M &            & 35.31 & 0.947 \\
    CST-M & \checkmark & \baseline{\textbf{35.42}} & \baseline{\textbf{0.948}} \\
    \hline
    SAUNet &            & 34.84 & 0.946 \\
    SAUNet & \checkmark & \baseline{\textbf{35.00}} & \baseline{\textbf{0.944}} \\
\end{tabular}
\caption{\textbf{Ablation study} of different HSI methods applying Range-Null space decomposition.}
\label{tab:ablation:different}
\end{table}

\begin{table}[!h]
\centering
\smalltablestyle{4pt}{1.05}
\begin{tabular}{l* cc|cc|cc}
    & \multicolumn{2}{c}{\footnotesize original} & \multicolumn{2}{c}{\footnotesize finetune + RND} & \multicolumn{2}{c}{\footnotesize PnP RND} \\
    method & PSNR & SSIM & PSNR & SSIM & PSNR & SSIM \\
    \shline
    TSA-Net~\cite{meng2020end} & 32.99 & 0.923 & 33.12 & 0.922 & 33.09 & 0.921 \\
    MST-S~\cite{mst}           & 34.26 & 0.935 & 34.63 & 0.940 & 34.43 & 0.937 \\
    MST-M~\cite{mst}           & 34.94 & 0.943 & 35.32 & 0.950 & 35.04 & 0.947 \\
    CST-S~\cite{cai2022coarse} & 34.71 & 0.940 & 34.85 & 0.943 & 34.84 & 0.943 \\
    CST-M~\cite{cai2022coarse} & 35.31 & 0.947 & 35.42 & 0.950 & 35.42 & 0.950 \\
\end{tabular}
\caption{\textbf{Fine-tuning and Plug-and Play with RND}.}
\label{tab:ablation:pnp_finetune}
\end{table}
\subsection{Generalization for RND-SCI Framework}
Our RND-SCI framework is a generic HSI reconstruction method and can be used in applications other than different types of methods and size models. To prove it, we apply RND-SCI to MST-S, MST-M, CST-S, CST-M, and SAUNet and train them from scratch. As shown in Table \ref{tab:ablation:different}. All methods model has improved at least 0.11 dB. It displays the generalization of RND-SCI for the HSI reconstruction task.

Moreover, fine-tune existing models with it or insert a pre-trained model directly into the RND-SCI framework to get better results. We finetune TSA-Net, MST-S, MST-M, CST-S, and CST-M at the beginning learning rate of $1\times 10^{-5}$, warm-up about 1000 steps, and for 50 epochs. The results are displayed in Table \ref{tab:ablation:pnp_finetune} indicates the diversity of RND-SCI implementations and also shows the potential of this decomposition method to solve the compressive spectral imaging inverse problem. 

\section{Conclusion and Future Work}
In this paper, we proposed a simple and flexible framework, RND-SCI, for snapshot compressive imaging. The proposed framework is based on the theory of the range-null space decomposition, which provides a new perspective on modeling data consistency (range-space term) and regularization (null-space term). The data-fidelity term can be obtained directly by a linear projection operation without learning. While the image prior term generates a solution that conforms to a real constraint through a conditional generation network, and then passes through a null-space projection operation. Finally, the two components are aggregated to derive the ultimate reconstruction outcome.

To achieve fast and simple reconstruction, we derive a simple projection operation approach inspired by the principles of CASSI and apply it to an enhanced version of SAUNet, achieving the 91 frames per second reconstruction speed and favorable accuracy. Moreover, we launch extensive experiments to investigate the framework robustness of the modulated mask, different neural networks, and different implementations. Compared with the original model methods, there is evident promotion.

In further work, we hope to take the unified method to extend more similar problems about the hyperspectral image field task, such as hyperspectral image restoration and hyperspectral image super-resolution, to improve the model performance and interpretation, and even develop a unified architecture to solve the set of similar hyperspectral imaging inverse problems on the basis of the theory.

\appendix

\section*{Appendix}
The supplementary material is organized as follows:

\begin{itemize}
\item Section \ref{sec1}: real HSI reconstruction for RND-SAUNet. 
\item Section \ref{sec2}: more comparisons for RND-SAUNet.
\item Section \ref{sec3}: parameter analysis for RND-SAUNet.
\item Section \ref{sec4}: limitations and social impacts.
\end{itemize}

\section{Real HSI Reconstruction for RND-SAUNet} \label{sec1}
\begin{figure*}[t]
    \centering
    \includegraphics[width=\linewidth,scale=1.00]{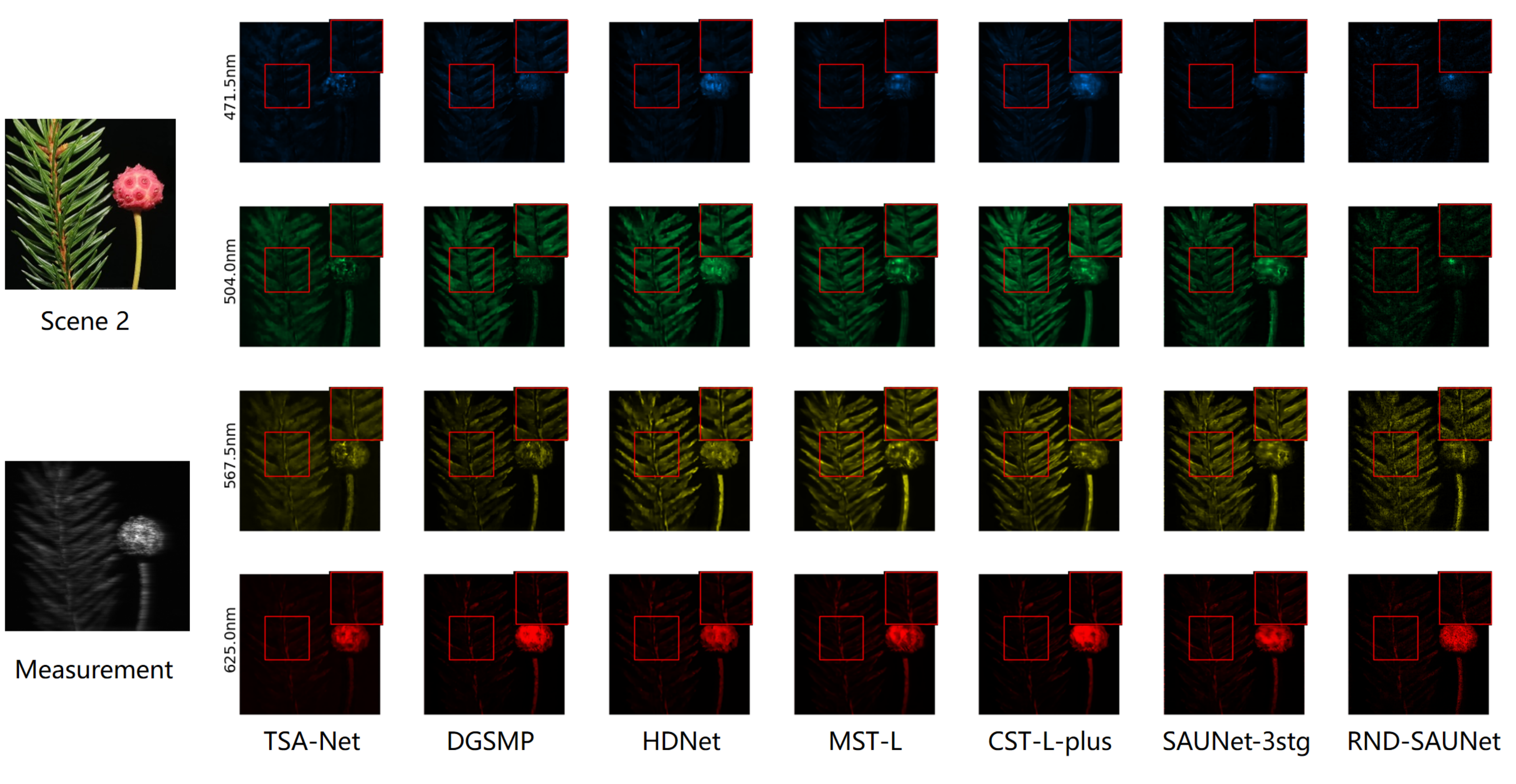}
    \caption{Reconstructed real HSI comparisons on Scene 2 with 4 out of 28
spectral channels. Six SOTA less time-consuming methods and RND-SAUNet are included. Zoom in for a better view.}
    \label{fig:fig1}
\end{figure*}

\begin{figure*}[t]
    \centering
    \includegraphics[width=\linewidth,scale=1.00]{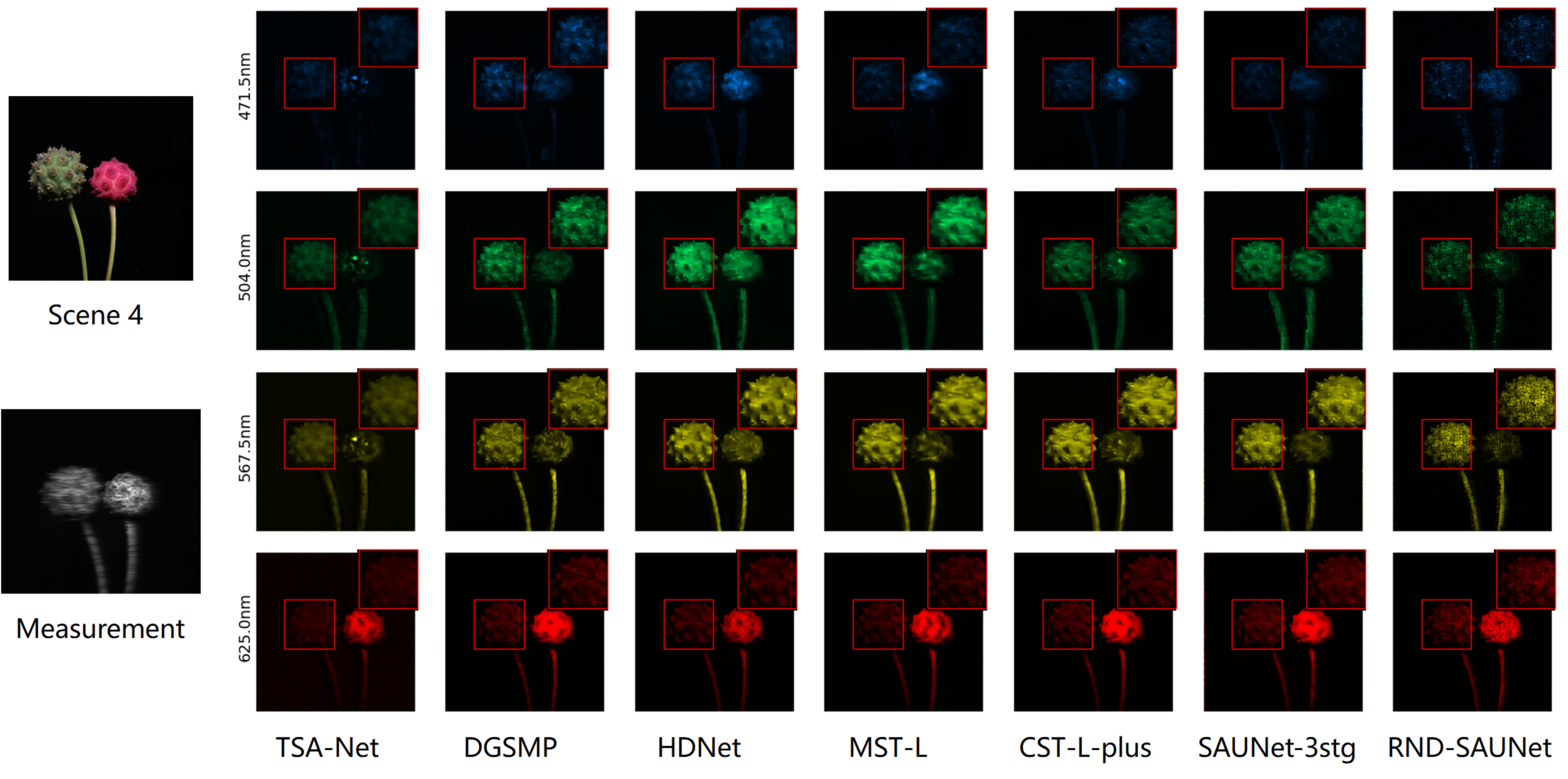}
    \caption{Reconstructed real HSI comparisons on Scene 4 with 4 out of 28
spectral channels. Six SOTA less time-consuming methods and RND-SAUNet are included. Zoom in for a better view.}
    \label{fig:fig2}
\end{figure*}

\paragraph{Implementation.} Following TSA-Net settings. we re-train RND-SAUNet with a real mask on the KAIST and CAVE datasets jointly, which our paper has mentioned, while the training samples are also injected with 11-bit shot noise for simulating the real imaging situations.
\paragraph{Real HSI reconstruction for RND-SAUNet.}
Figure \ref{fig:fig1} and Figure \ref{fig:fig2} show the visual comparisons between our RND-SAUNet and other less time-consuming methods. although our methods can reconstruct the structure of the scene and some details, suffering from the lack of design in imaging noise, the simple image plus operator can not remove the noise, which results in the HSI reconstruction having some hollows.

\section{More Comparisons for RND-SAUNet} \label{sec2}
\paragraph{Training memory.} Table \ref{tab:mem} shows the memory required to train the HSI algorithms. For a fair comparison,  We set the batch size as 5 for all algorithms, except DGSMP for batch size as 2. Compared to other methods, RND-SAUNet is memory-friendly, which only costs 6.50 GB of memory during training.  
\begin{table*}[!h]
    \centering
    \resizebox{1.0\linewidth}{!}{
        \begin{tabular}{c|ccccccccc}
           Model
           & TSA-Net
           & DGSMP
           & GAP-Net
           & ADMM-Net
           & MST-S
           & CST-S
           & HDNet
           & SAUNet-1stg
           & \cellcolor{Highlight}RND-SAUNet(Ours)\\ \shline
            Training Memory(GB)
            &  11.20
            &  13.79
            &  8.17
            &  8.78
            &  9.19
            &  8.14
            &  7.83
            &  7.48
            & \baseline{\textbf{6.50}}
        \end{tabular}
    }
    \caption{\textbf{Comparisons between RND-SAUNet and 8 SOTA less time-consuming method in training memory}. We set the batch-size as 5 for all algorithms, expect DGSMP for 2.}
    \label{tab:mem}
\end{table*}

\begin{figure*}[t]
    \centering
    \includegraphics[width=\linewidth,scale=1.00]{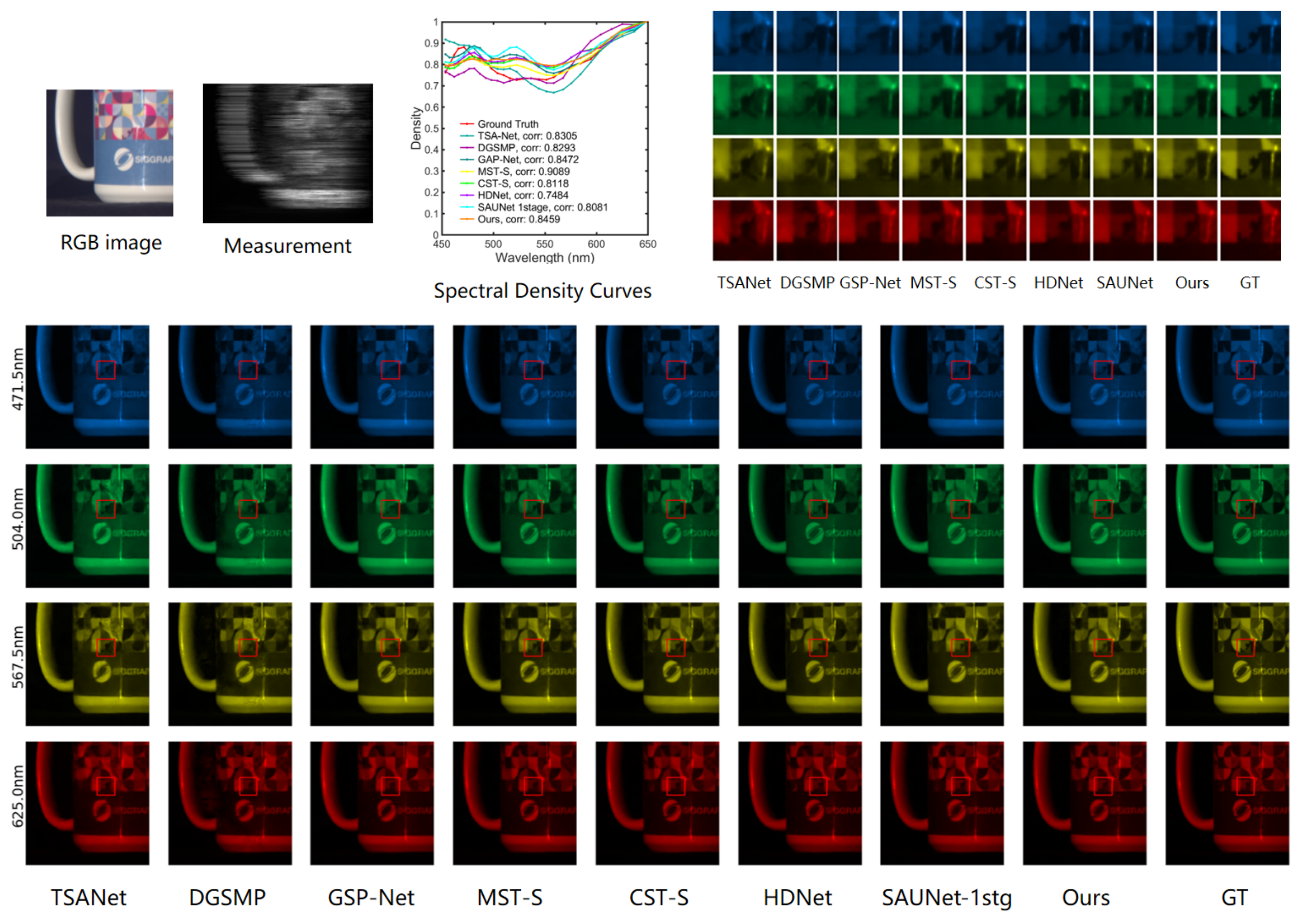}
    \caption{Reconstructed real HSI comparisons on Scene 5 with 4 out of 28
spectral channels. Seven SOTA less time-consuming methods and RND-SAUNet are included. Zoom in for a better view.}
    \label{fig:img1}
\end{figure*}

\begin{figure*}[t]
    \centering
    \includegraphics[width=\linewidth,scale=1.00]{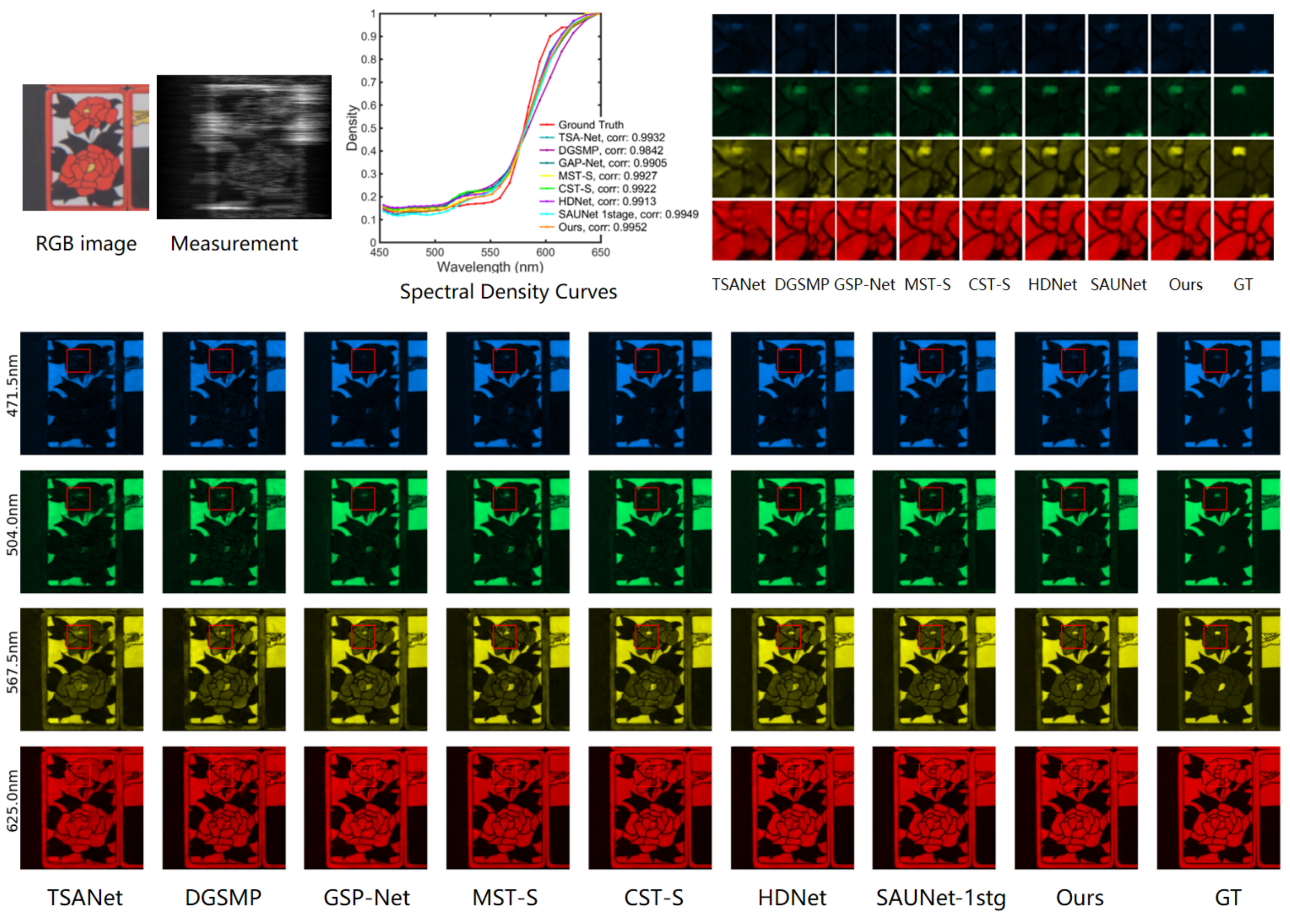}
    \caption{Reconstructed real HSI comparisons on Scene 7 with 4 out of 28
spectral channels. Seven SOTA less time-consuming methods and RND-SAUNet are included. Zoom in for a better view.}
    \label{fig:img2}
\end{figure*}

\paragraph{Other visual comparisons.} Figure \ref{fig:img1} and \ref{fig:img2} show the reconstructed simulation HSI comparisons of Scene 5, 7, and 10 with 4 out of 28 spectral channels. Compared to Seven less time-consuming algorithms. Please zoom in for a better view. As can be seen from the reconstructed HSI, our method RND-SAUNet can recover high-frequency HSI contents and structural textures, resulting in producing perceptually-pleasing images.

\section{Parameter Analysis for RND-SAUNet} \label{sec3}
To further investigate the influence of the changes of the initial network and cropping operator, we compare the two hyper-parameters value $\alpha$ and $\beta$ between the SAUNet-1stg and RND-SAUNet. As shown in the table \ref{tab:hyper}, the $\alpha$, the value of $\alpha$ increases while $\beta$ decreases, which may indicate a better recovery of the image via these improvements.

\begin{table}[!h]
\centering
\begin{tabular}{c|cc}
    Method & $\alpha$ & $\beta$\\
    \shline
    SAUNet-1stg & 325.08 & $0.6 \times 10^-3$  \\
    RND-SAUNet  & 629.91 & $1.0 \times 10^-6$  \\
\end{tabular}
\caption{\textbf{The hyper-parameter of SAUNet-1stg and RND-SAUNet}.}
\label{tab:hyper}
\end{table}

\section{Limitations and Social Impacts.} \label{sec4}

There remain some limitations that deserve further study.

\begin{itemize}
\item Though RND-SAUNet achieves the fastest results and gets favorable reconstruction performance among the less time-consuming methods, it is still limited in accuracy compared to the high-precision methods, such as DAUHST.

\item RND-SCI needs a certain coded aperture mask. It may lead to the model not adapting to the change in the CASSI system.
\item RND-SCI does not design to remove imaging noise of the real scene.

\end{itemize}

Hyperspectral compressive imaging is one of the core tasks in snapshot compressive imaging (SCI) and has been developed for decades. It has been widely used for object detection~\cite{yan2021object}, medical imaging~\cite{lu2014medical}, remote sensing~\cite{van2012multi}, and so on. Our algorithm is aimed to improve the speed of reconstruction over all existing other SOTA methods (RND-SAUNet) and provide a unified and easy-to-implement framework (RND-SCI) for a better reconstruction performance. Until now, HSI reconstruction has had no negative social impact, but our study involves GPU resources for training the models, which will result in $\mathrm{CO_2}$ emissions.

{\small
\bibliographystyle{ieee_fullname}
\bibliography{rndsci}
}

\end{document}